\documentclass[conference]{IEEEtran}
\IEEEoverridecommandlockouts

\usepackage{cite}
\usepackage{amsmath,amssymb,amsfonts}
\usepackage{algorithmic}

\usepackage{graphicx}
\usepackage{textcomp}
\usepackage{xcolor}
\def\BibTeX{{\rm B\kern-.05em{\sc i\kern-.025em b}\kern-.08em
    T\kern-.1667em\lower.7ex\hbox{E}\kern-.125emX}}
    
\usepackage{amsmath,amsfonts}

\usepackage{times}
\usepackage{url}
\usepackage{hyperref}
\usepackage[utf8]{inputenc}
\usepackage[small]{caption}
\usepackage{graphicx}
\usepackage{amsmath}
\usepackage{amsthm}
\usepackage{booktabs}
\usepackage{enumitem}

\usepackage{algorithm}
\usepackage{algorithmic}

\usepackage{mathrsfs} 

\usepackage{nicefrac}       
\usepackage{microtype}      
\usepackage{graphicx}
\usepackage{graphics}
\usepackage{rotating}
\usepackage{mathrsfs} 
\usepackage{array,multirow,multicol}
\usepackage{lscape} 
\usepackage{array}
\usepackage{multicol}
\usepackage{xcolor}
\usepackage{colortbl}
\usepackage{todonotes}
\graphicspath{{figs/}}
\usepackage[super]{nth}

\usepackage{algorithmic}
\usepackage{array}
\usepackage[caption=false,font=normalsize,labelfont=sf,textfont=sf]{subfig}
\usepackage{textcomp}
\usepackage{stfloats}
\usepackage{url}
\usepackage{verbatim}
\usepackage{graphicx}

\usepackage{authblk}
\usepackage{soul}
\soulregister\cite7
\soulregister\ref7
\soulregister\pageref7

\newtheorem{mydef}{Definition}
\newtheorem{mynotion}{Notation}
\newtheorem{assumption}{Assumption}
    
\begin{document}

\title{Improve ROI with Causal Learning and Conformal Prediction}



\author{
Meng Ai, Zhuo Chen, Jibin Wang, Jing Shang, Tao Tao, Zhen Li
\\
China Mobile Information Technology Co., Ltd, Beijing, China
\\
\{aimeng, chenzhuoit, wangjibin, shangjing, taotao, lizhenit\}@chinamobile.com
\\
}

\maketitle

\begin{abstract}

In the commercial sphere, such as operations and maintenance, advertising, and marketing recommendations, intelligent decision-making utilizing data mining and neural network technologies is crucial, especially in resource allocation to optimize ROI. This study delves into the Cost-aware Binary Treatment Assignment Problem (C-BTAP) across different industries, with a focus on the state-of-the-art Direct ROI Prediction (DRP) method. However, the DRP model confronts issues like covariate shift and insufficient training data, hindering its real-world effectiveness. Addressing these challenges is essential for ensuring dependable and robust predictions in varied operational contexts.

This paper presents a robust Direct ROI Prediction (rDRP) method, designed to address challenges in real-world deployment of neural network-based uplift models, particularly under conditions of covariate shift and insufficient training data. The rDRP method, enhancing the standard DRP model, does not alter the model's structure or require retraining. It utilizes conformal prediction and Monte Carlo dropout for interval estimation, adapting to model uncertainty and data distribution shifts. A heuristic calibration method, inspired by a Kaggle competition, combines point and interval estimates. The effectiveness of these approaches is validated through offline tests and online A/B tests in various settings, demonstrating significant improvements in target rewards compared to the state-of-the-art method.

\end{abstract}


\section{Introduction}
\label{sec:introduction}

In a wide range of commercial activities, intelligent decision-making based on data mining and neural network technologies is playing an increasingly important role. One crucial aspect of this intelligent decision-making is figuring out how to allocate limited resources in order to maximize returns, essentially boiling down to the problem of estimating ROI. For instance, in the field of operations and maintenance, how to allocate machine resources and computational power to maximize the revenue of supported businesses \cite{jiang2020dcaf}; in the advertising sector, how to distribute an advertiser's total budget reasonably to maximize the revenue from their products \cite{jin2018real}; and in the realms of recommendation and marketing, how to allocate suitable coupons, discounts, and coins as incentives to users in order to maximize platform user retention, GMV, etc \cite{Hua2021Markdowns,Zhang2021BCORLE,zhou2023direct,Ai2022LBCF,Moraes2023uplift,ziang2023an}.

In causal inference, actions such as adjusting the computational power for a specific business operation, modulating the cost of a particular advertisement, and offering incentives of varying value, as mentioned in the above examples, are collectively referred to as treatment. In this study, we primarily consider a simple yet very common scenario: the presence or absence of treatment. Therefore, the decision problems involving resource allocation mentioned above can all be defined as Cost-aware Binary Treatment Assignment Problem (C-BTAP) \cite{zhou2023direct,Ai2022LBCF,Du2019Improve}. 

\begin{figure}
\centering
\subfloat[]{\includegraphics[width=0.25\textwidth]{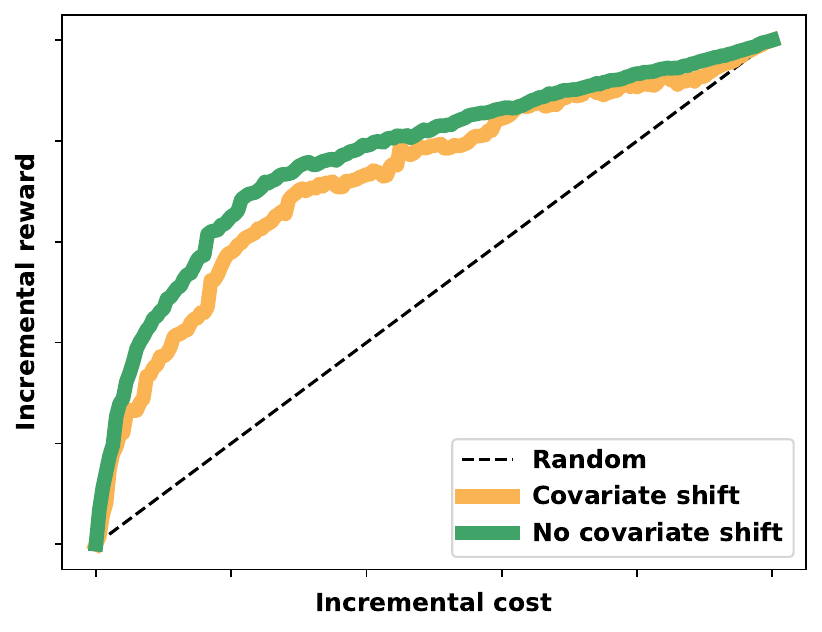}%
\label{fig:mape}}
\subfloat[]{\includegraphics[width=0.25\textwidth]{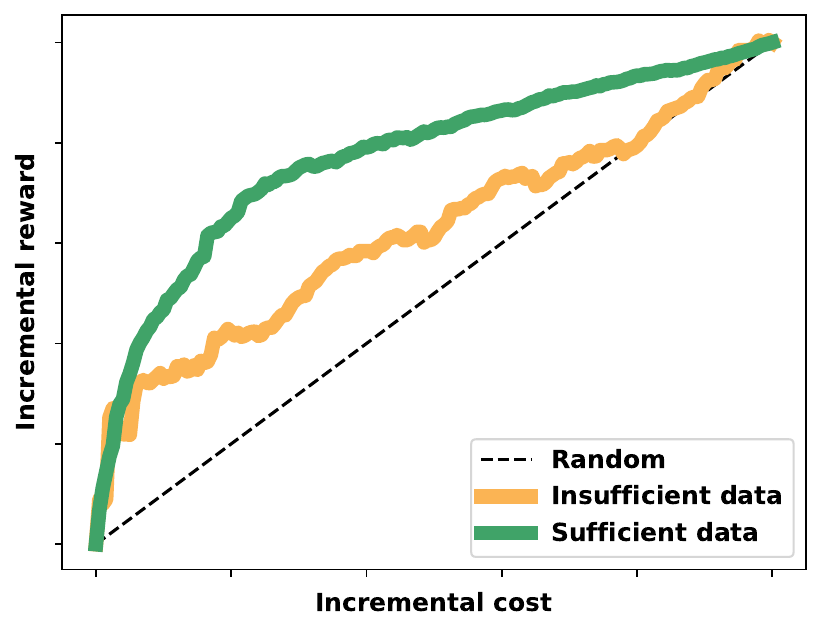}%
\label{fig:mape2}}
\caption{Two limitations lead to a decline in DRP's performance on the test set. A larger area under the curve indicates better performance. (a) Covariate shift. (b) Insufficient data.}
\label{fig:limits}
\vspace{-15pt} 
\end{figure}

Combining heterogeneous causal effect estimation (a.k.a. uplift model) and binary operations research (OR) optimization techniques, C-BTAP has been studied in many academic papers in recent years ~\cite{zhou2023direct, Ai2022LBCF,Zhao2019Unified,Du2019Improve,ziang2023an}. Three popular methods have been proposed for tackling C-BTAP: 1) Two-Phase Methods (TPM). First, TPM utilized uplift models, such as meta-learners~\cite{kunzel2019metalearners,Nie2021Quasi}, causal forests~\cite{Wager2018Estimation,Athey2019Generalized,Zhao2017Uplift,Ai2022LBCF}, or neural network based representation learning~\cite{Johansson2016Learning,shalit2017estimating,Yao2018Representation} approaches, to predict the revenue lift and cost lift, respectively. Then, a calculation is performed by dividing the revenue uplift prediction by the cost uplift prediction. However, combination of revenue uplift model and cost uplift model may cause an enlargement of model errors due to the mathematical operations during combination; 2) For the method of Direct Rank (DR), a loss function aimed at ranking individuals' ROI is created, as noted in ~\cite{Du2019Improve}. However,  \cite{zhou2023direct} demonstrate that achieving accurate ranking is not possible when the loss function fully converges because the loss function is not convex, which is also detailed in Appendix E of \cite{zhou2023direct}; 3) based on our research of the published literature, the Direct ROI Prediction (DRP) method \cite{zhou2023direct}, presented at AAAI 2023, remains the state-of-the-art (SOTA) for C-BTAP so far. DRP designs a convex loss function for neural networks to guarantee an unbiased estimation of ROI of individuals when the loss converges.

However, what if some factors in the real-world deployment environment cause the loss function to not converge effectively? Next, we identify two limitations of the DRP model in its practical application in the industry, which lead to a decline in the DRP's performance on the test set as in Fig. \ref{fig:limits}. The first limitation is covariate shift. Covariate shift refers to the inconsistency in the distribution between a model's training and test sets. This can lead to a situation where, even though the DRP's loss function converges smoothly on the training set, it may be far from the convergence point on the test set. For example, the training set was collected on workdays, but the model is deployed during holidays. The ninety percent of of users in the training set might be office workers, while at least half of the users in the test set could be urban tourists. This is a covariate shift in demographic characteristic. The second limitation is insufficient training samples. In scenarios with insufficient samples, it becomes more challenging to configure and adjust aspects such as the learning rate, initial weights, and batch size for neural networks. This makes it difficult for DRP to converge on the training samples, not to mention on the test samples during real-world deployment, even in the absence of covariate shift. For instance, uplift models like DRP require training data collected through Randomized Controlled Trials (RCT). However, RCT experiments themselves can impact user experience or platform revenue. As a result, RCT usually occupy only a small fraction of online traffic, such as 0.1\%, and cannot be too long. This leads to insufficient training data for uplift models like DRP. 

To mitigate the impact of the aforementioned two limitations on real-world deployment applications, based on the DRP model \cite{zhou2023direct}, a robust Direct ROI Prediction (rDRP) method is proposed in this study.
For the neural network-based uplift models that do not have the actual label, there is not too much research in this area. For instance, \cite{sun2023robustnessenhanced} presents an example of adversarial robustness in uplift models. However, our research aligns more closely with distributional robustness. Therefore, we propose the rDRP method, aiming to explore the domain of robustness in ROI prediction.
rDRP solely utilizes the information from prediction intervals to perform post-processing calibration on the point estimates of DRP. Consequently, the rDRP method needs to address two issues: first, estimating the prediction intervals for ROI models like DRP; and second, conducting the post-processing calibration.

For the first issue, existing uplift modeling employs methods like causal forest \cite{Wager2018Estimation}, which uses infinitesimal jackknife for Conditional Average Treatment Effects (CATE) variance estimation, X-learner \cite{kunzel2019metalearners} applies bootstrap for the same purpose and Bayesian Additive Regression Trees (BART) \cite{chipman2010bart, hill2011bayesian, green2012modeling, hahn2020bayesian} for credible intervals in causal inference. A notable approach \cite{lei2021conformal} applies conformal inference for valid uncertainty estimates in individual treatment effects. However, these techniques for uncertainty quantification (UQ) in uplift modeling are not directly applicable to ROI UQ, as they require separate predictions for revenue and cost uplift, whereas ROI prediction intervals involve a more complex interplay between these two factors. Besides, methods like Bootstrap or Model Ensemble require retraining multiple models, resulting in lower efficiency. Therefore, we propose an direct interval estimate for ROI, not requiring combining two uplift models nor retraining multiple models. 

Our ROI interval estimation method is based on conformal prediction for three reasons, with the first two addressing the aforementioned two limitations. The first reason: conformal prediction does not require a specific data distribution for the training set, making it suitable for scenarios with covariate shift between the training and test sets. It only requires that the calibration set and test set have consistent distributions. Therefore, we can conduct a one or two days RCT to collect calibration set data (as a comparison, training datasets typically require at least 2-3 weeks) before the actual deployment of the uplift model to ensure that the calibration set and test set are as consistently distributed as possible. The second reason: the interval predictions from conformal prediction can adapt well to the model’s inherent uncertainty due to insufficient training. For instance, if the model shows greater uncertainty on certain data points due to insufficient training data, the conformal prediction will produce wider prediction intervals. This feature helps us determine which point estimates might be less reliable when the DRP model has insufficient training data. The third reason: conformal prediction does not require altering the neural network structure of the original DRP model, nor does it necessitate retraining the model.

The key aspect of the conformal prediction method is how to obtain the conformal score. Inspired by \textit{Conformalizing Scalar Uncertainty Estimates} in \cite{angelopoulos2021gentle}, we define the Conformal Score as a combination of three key elements, specifically including the point estimate from the DRP, the standard deviation (std) of the DRP's point estimate, and the point estimate of the DRP's loss function at the convergence point. The point estimate for the DRP can be obtained from the original DRP inference; during DRP inference, we add a Monte Carlo (MC) dropout \cite{gal2016dropout} layer to calculate the std of the DRP point estimate. Finally, leveraging the convex nature of the DRP loss function, we employ binary search on the calibration set to find the convergence point.

For the second issue, according to our research, there are not many methods on how to use prediction intervals to calibrate point estimates in the context of uplift models, such as those used for ROI prediction. Inspired by a Kaggle competition, we propose a heuristic method for calibration. \textit{The 2018 M4 Forecasting Competition} by Kaggle was the first M-Competition to elicit
prediction intervals in addition to point estimates. \cite{Grushka-Cockayne2020-qv} summarized six effective interval aggregation methods used in the competition. Although the goals are somewhat different, the competition's focus is on time series, while our focus is on ROI ranking sequence. The heuristic method we propose involves calibrating point estimates with interval estimates in several forms, such as weighting the point estimates using the upper bound of the prediction interval or the width of the prediction interval. Then, validation is performed on the calibration set to select which form of calibration is best.

Finally, to validate the effectiveness of our approaches, offline tests and online A/B Tests are conducted in four different settings, based on whether there are sufficient samples and whether there is covariant shift between the calibration/test and training sets. In the offline simulations, we use three widely recognized and evaluated real-world datasets for uplift related models. The evaluation metrics and online A/B tests show that our models and algorithms achieve significant improvement compared with state-of-the-art. To facilitate application in real-world scenarios, we also discuss in detail the limitations of the method we propose.

\section{Related Work}
Our work relates to uplift model and its application in resource allocation. Besides, uncertainty quantification and robust optimization are also related. Thus, we review the state-of-the-art methods from these aspects in this section. 


\subsection{Resource Allocation} The methods for solving the resource allocation problem can be divided into three categories. 

The first category is the two-phase methods. In the first phase, uplift models, such as meta-learners~\cite{kunzel2019metalearners,Nie2021Quasi}, causal forests~\cite{Wager2018Estimation,Athey2019Generalized,Zhao2017Uplift,Ai2022LBCF}, and representation learning~\cite{Johansson2016Learning,shalit2017estimating,Yao2018Representation} approaches, are employed to predict incremental responses to different treatments. In the second phase, Lagrangian duality, a powerful algorithm for decision-making issues, is commonly used, finding applications in domains like marketing budget allocation~\cite{Du2019Improve,Ai2022LBCF,Zhao2019Unified} and formulating optimal bidding strategies in online advertising~\cite{Hao2020Dynamic}. However, ROI of individuals is a composite object and all these works predict it by the combination of benefit uplift model and cost uplift model, which may cause an enlargement of model errors due to the mathematical operations during combination.

The second category is Direct Learning method. This method focuses on learning a treatment assignment policy directly, rather than treatment effects, thus circumventing the need to merge ML and OR techniques as discussed in \cite{Xiao2019Model,Zhang2021BCORLE}. A framework for policy learning using observational data was introduced, based on the doubly robust estimator \cite{Athey2021Policy}. This concept was further expanded for multi-action policy learning in \cite{Zhou2022Offline}. Nevertheless, these approaches incorporate resource constraints into the reward function via a Lagrangian multiplier, which might necessitate frequent modifications to the model in response to changes in the Lagrangian multiplier.


The third category is Decision-focused learning. This approach concentrates on training the model parameters focused on the downstream optimization task, rather than solely on prediction accuracy, as explored in \cite{Bryan2019DFL,Adam2022SPO,Shah2022LODL,Mandi2022Decision}. However, these studies assume that the feasible region for the decision variables is stable and definitively known. \cite{Du2019Improve,zou2020heterogeneous,goldenberg2020free} propose to directly learn the ratios between revenues and costs in the binary treatment setting, and treatments could be first applied to users with higher scores. However, as \cite{zhou2023direct} proves, the proposed loss in~\cite{Du2019Improve,zou2020heterogeneous} cannot achieve the correct rank when
the loss converges. The Direct ROI Prediction (DRP) method \cite{zhou2023direct} is also categorized under this approach. It avoids the disadvantages of combing two uplift models by using a single model to directly predict ROI. However, the robustness of this method during actual deployment needs to be improved, particularly in cases of insufficient samples or covariate shifts.

\subsection{Uncertainty Quantification}


Recently, uncertainty quantification (UQ) has become a key focus in research, with wide-ranging applications in real-world scenarios, as discussed in \cite{abdar2021review}, \cite{gawlikowski2021survey}. This field generally categorizes uncertainty into aleatoric and epistemic types. Aleatoric uncertainty arises from data's randomness and is quantifiable, often using predictive means and variances with a negative log-Gaussian likelihood \cite{kendall2017uncertainties}. Epistemic uncertainty, caused by data scarcity, is reducible and commonly estimated with Bayesian Neural Networks (BNNs) \cite{jospin2020hands}. BNNs, which increase model parameters and require computing the Kullback-Leibler (KL) divergence \cite{blundell2015weight}, can be complex. A simpler alternative is the Monte Carlo (MC) dropout \cite{gal2016dropout}, using dropout at testing to approximate Bayesian analysis, differing from standard dropout \cite{srivastava2014dropout}. Conformal prediction (CP) \cite{cpbook} assesses a model's confidence in predictions through validity, ensuring accuracy within a set confidence level. It changes point predictions to intervals or sets, indicating confidence levels, where larger intervals signal more uncertainty. CP divides test data into calibration and test sets, with calibration used for setting confidence thresholds. This approach is versatile, applied in fields like image classification \cite{raps} and regression \cite{cqr}.

In uplift modeling, various methodologies offer ways to quantify uncertainty. For instance, causal forest \cite{Wager2018Estimation} employs infinitesimal jackknife \cite{efron2014estimation, wager2014confidence} for CATE variance estimation, assuring asymptotic coverage under certain assumptions. Similarly, X-learner \cite{kunzel2019metalearners} uses bootstrap for this purpose. Bayesian Additive Regression Trees (BART), originally a Bayesian machine learning tool \cite{chipman2010bart}, were adapted for causal inference \cite{hill2011bayesian, green2012modeling, hahn2020bayesian}, demonstrating superior accuracy and coverage \cite{dorie2017aciccomp2016, dorie2019automated}, and providing credible intervals for CATE. \cite{lei2021conformal} introduces a method that applies conformal inference techniques to provide reliable and statistically valid uncertainty estimates for individual treatment effects and counterfactual predictions in causal analysis. However, for the UQ of ROI, these methods can only predict the UQ of revenue uplift or cost uplift separately. The prediction interval for ROI is not merely a matter of dividing the prediction interval of revenue uplift by that of cost uplift.

\subsection{Robust Optimization}


Robust optimization is a key area of interest in the machine learning field, primarily focusing on two popular forms: distributional robustness and adversarial robustness. Distributional robustness, as discussed in \cite{rahimian2019distributionally}, addresses the issue of discrepancies between training and testing distributions. Adversarial robustness, detailed in \cite{bai2021recent}, ensures model stability even when inputs are slightly altered. \cite{sun2023robustnessenhanced} presents an example of adversarial robustness in uplift models. However, our research aligns more closely with distributional robustness. There's a notable gap in existing literature regarding distributional robustness in ROI prediction models, especially since such models, like uplift models, often lack actual labels.

\section{PRELIMINARIES}
\label{sec:PRELIMINARIES}

In this section, we first introduce notations, definitions and assumptions used in this paper. Next, we define the resource assignment problem and elucidate the objective of this study.

\subsection{Notations and Definitions}\label{nota:def}
In this study, we adopt the potential outcome framework~\cite{sekhon2008neyman} in causal inference to formulate this problem. 

\begin{mynotion}[Feature, Outcome, Treatment]
Let $X \in \mathbb{R}^d$ denote the feature vector and $x_i$ its realization. 
Despite the incremental revenue, marketing actions can also incur significant costs. Let $Y^r,Y^c$ denote the revenue outcome and the cost outcome respectively, and $y^r_i, y^c_i$ its realization. 
Denote the treatment by $T \in \{0,1\}$ and its realization by $t_i$. 
\end{mynotion}

\begin{mynotion}[Potential Outcome]
Let $(Y^r(1), Y^r(0))$ and $(Y^c(1), Y^c(0))$ be the corresponding potential outcome when the individual receives the treatment or not. 
\end{mynotion}

\begin{mydef}[CATE]
Define $\tau^r(x_i), \tau^c(x_i)$ as the Conditional Average Treatment Effect (CATE), a.k.a. uplift, which can be calculated by 
$$\tau^*(x_i) = E[Y^*(1)-Y^*(0)|X=x_i], * \in \{r,c\}.$$

\end{mydef}

\begin{mydef}[ROI]
Define $\tau^r(x_i)/\tau^c(x_i)$ as Return on Investment (ROI) of individual $i$, denoted by $roi_i$. 
\end{mydef}

\begin{mydef}[C-BTAP]
\label{def:btap}
The Cost-aware Binary Treatment Assignment Problem (C-BTAP) involves assigning binary treatments to a subset of individuals to maximize total revenue on a platform, ensuring that the additional costs incurred stay within a fixed budget limit $B$.

Besides, we make the following assumptions in this study.

\begin{assumption}
\label{assu:diff_within3}
RCT Sample. Consider a dataset of size $N$ derived from Random Control Trials (RCT), where the i-th entry is represented as $(x_i, t_i, y^r_i, y^c_i)$. The number of samples receiving the treatment is denoted by $N_1$, while those not receiving it are denoted by $N_0$.
\end{assumption}

\begin{assumption}
\label{assu:sutva}
SUTVA. It stands for the Stable Unit Treatment Value Assumption. This assumption, crucial in the context of causal inference as highlighted by~\cite{sekhon2008neyman}, posits that the potential outcome for any individual unit remains uninfluenced by the treatment assignments given to other units. Based on the random nature of RCT, this assumption will hold in this study. 
\end{assumption}

\begin{assumption}
\label{assu:diff_within2}
ROI's Scope. By scaling and truncating $Y^r$ or $Y^c$, we constrain ROI within the range of $(0,1)$. This limitation reduces the risk of overfitting, particularly important as the division in ROI calculations can lead to high variance, especially when $\tau^c(x_i)$ is low.
\end{assumption}

\begin{assumption}
\label{assu:diff_within}
Positive Treatment Effect.  Given that the majority of marketing interventions positively influence an individual's response and also entail a positive cost, we posit that $\tau^r(x_i) > 0$ and $\tau^c(x_i) > 0$.
\end{assumption}

\end{mydef}

\subsection{Problem Formulation}

In a typical marketing situation, a selection is made from a group of $M$ individuals to receive a marketing initiative. The decision variables are denoted as $z_i \in \{0,1\}$. Thus, the C-BTAP is framed as the integer programming problem expressed in equation~\eqref{BTAP}.

\begin{align}
\max& \sum_i z_i \tau^r(x_i) \label{BTAP}\\
s.t.& \sum_i z_i \tau^c(x_i) \le B& \nonumber\\
& z_i \in \{0,1\}, \forall i.& \nonumber
\end{align}


The C-BTAP mirrors the 0/1 knapsack problem, known to be NP-Hard. Fortunately, the straightforward greedy Algorithm~\ref{alg:BTAG} delivers impressive results. Its approximation ratio is $\rho \ge 1 - \frac{\max_i \tau(x_i)}{\mathrm{OPT}}$, with $\mathrm{OPT}$ representing the optimal solution to Equation~\eqref{BTAP}.

\begin{algorithm}[tb]
\caption{Greedy Algorithm for C-BTAP}
\label{alg:BTAG}
\textbf{Input}: $(\tau^r(x), \tau^c(x), B, M)$\\
\textbf{Output}: C-BTAP's solutions.
\begin{algorithmic}[1] 
\STATE Sort individuals from set $M$ by their ROI, defined as $\frac{\tau^r(x_i)}{\tau^c(x_i)}$, from highest to lowest.
\STATE Allocate a binary treatment to each sorted individual until the budget $B$ is reached.

\end{algorithmic}
\end{algorithm}


\subsection{The Goal}

Algorithm \ref{alg:BTAG} establishes the connection between ROI prediction and C-BTAP. Thus, solving the C-BTAP problem essentially boils down to predicting ROI. Therefore, in this study, the methods proposed and the evaluation metric revolve around the prediction of ROI.

\section{METHOD}
\label{sec:method}
Firstly, we briefly give an overview of this part. In subsection \ref{sec:a}, we will review the Direct ROI Prediction (DRP) model, which remains the state-of-the-art (SOTA) model for C-BTAP so far. In subsection \ref{sec:b}, we point out two limitations of DRP. Then a robust DRP (rDRP) method is proposed to mitigate these limitations in subsection \ref{sec:c}. Subsection \ref{sec:d} performs a time complexity analysis of rDRP for practical use.


\subsection{DRP Model}
\label{sec:a}

Since our method rDRP is based on DRP, we first offer a concise overview of the DRP model, including its advantages and its loss function.

\textbf{DRP.} Proposed by \cite{zhou2023direct} in \textit{AAAI 2023}, DRP is a Direct ROI Prediction model for Cost-aware Binary Treatment Assignment Problem (BTAP). Based on our research of the published literature, the DRP remains the State-Of-The-Art (SOTA) model for C-BTAP so far.

\textbf{Advantage.} The ROI for individuals is typically predicted using a combination of different models. Common approaches for estimating $\tau^r(x_i)$ and $\tau^c(x_i)$ include employing meta-learners or causal forest models. However, using multiple models can amplify errors due to the calculations involved in their integration. To address this, DRP, a neural network-based approach, has been developed as a direct causal learning model for ROI prediction. This method has shown promising results in both theoretical and practical evaluations.



\textbf{Loss Function.} To train neural networks on the training set, DRP designed a loss function as follows.

\begin{align}
\min& L(s) = - [\frac{1}{N_1}\sum_{i|t_i=1} (y^r_i \ln \frac{\hat{roi_i}}{1-\hat{roi_i}} + y^c_i \ln (1 - \hat{roi_i})) - \nonumber\\ &\ \ \ \ \ \ \ \ \ \ \ \  \frac{1}{N_0}\sum_{i|t_i=0} (y^r_i \ln \frac{\hat{roi_i}}{1-\hat{roi_i}} + y^c_i \ln (1 - \hat{roi_i}))], \label{DRP} \\
s.t.& \ \ \ \ \ \ \ \ \ \ \ \ \ \ \ \ \hat{roi_i} = \sigma(\hat{s_i}) \in (0, 1), \forall i& \nonumber\\
& \ \ \ \ \ \ \ \ \ \ \ \ \ \ \ \ \hat{s_i} = \hbar(x_i) \in \mathbb{R}, \forall i.& \nonumber
\end{align}
where $\hat{roi_i}$ is the sigmoid transformation of $\hat{s_i}$ and $\hat{s_i}$ is the predicted point estimation of any neural network $\hbar(x_i)$. Based on the above four assumptions, \cite{zhou2023direct} proved in theory that the predicted $\hat{roi_i}$ is an unbiased point estimation when the loss Equation (\ref{DRP}) converges. 


\subsection{Limitations of DRP}
\label{sec:b}
Below, we identify two limitations of the DRP model in its practical application in the industry. These limitations might hinder DRP from appropriately obtaining unbiased point estimate ${roi^*}$ during inference on the test set, which lead to a decline in the DRP's performance as in Fig. \ref{fig:limits}.

\subsubsection{Covariate Shift}
\label{sec:limit1}

\begin{figure}[H]
    \centering
    \includegraphics[width=0.3\textwidth]{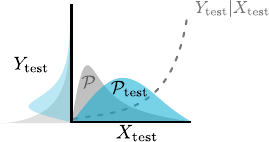}
    \caption{Covariate shift occurs as the distribution of $X_{\rm test}$ changes from $\mathcal{P}$ to $\mathcal{P}_{\rm test}$.}
    \label{fig:shift}
\end{figure}

Suppose the distribution of $X_{\rm train}$ is $\mathcal{P}$.
Covariate shift refers to the situation where the distribution of $X_{\rm test}$ changes from $\mathcal{P}$ to $\mathcal{P}_{\rm test}$, but the relationship between $X_{\rm test}$ and $Y_{\rm test}$, i.e. the distribution of $Y_{\rm test}|X_{\rm test}$, stays fixed. Fig. \ref{fig:shift} illustrates this point.

This problem is common in the real world.
For example,
\begin{itemize}
    \item You are attempting to gather training samples for the DRP model from an online RCT to predict the ROI of carpooling coupons. 
    Your model was trained on a dataset of 90\% office workers and 10\% tourists. Yet, during actual deployment, like on weekends, holidays, the ratio could shift to 50\%:50\%. This is a covariate shift in demographic characteristic.
\end{itemize}

\subsubsection{Insufficient Samples}
\begin{figure}[H]
    \centering
    \includegraphics[width=0.25\textwidth]{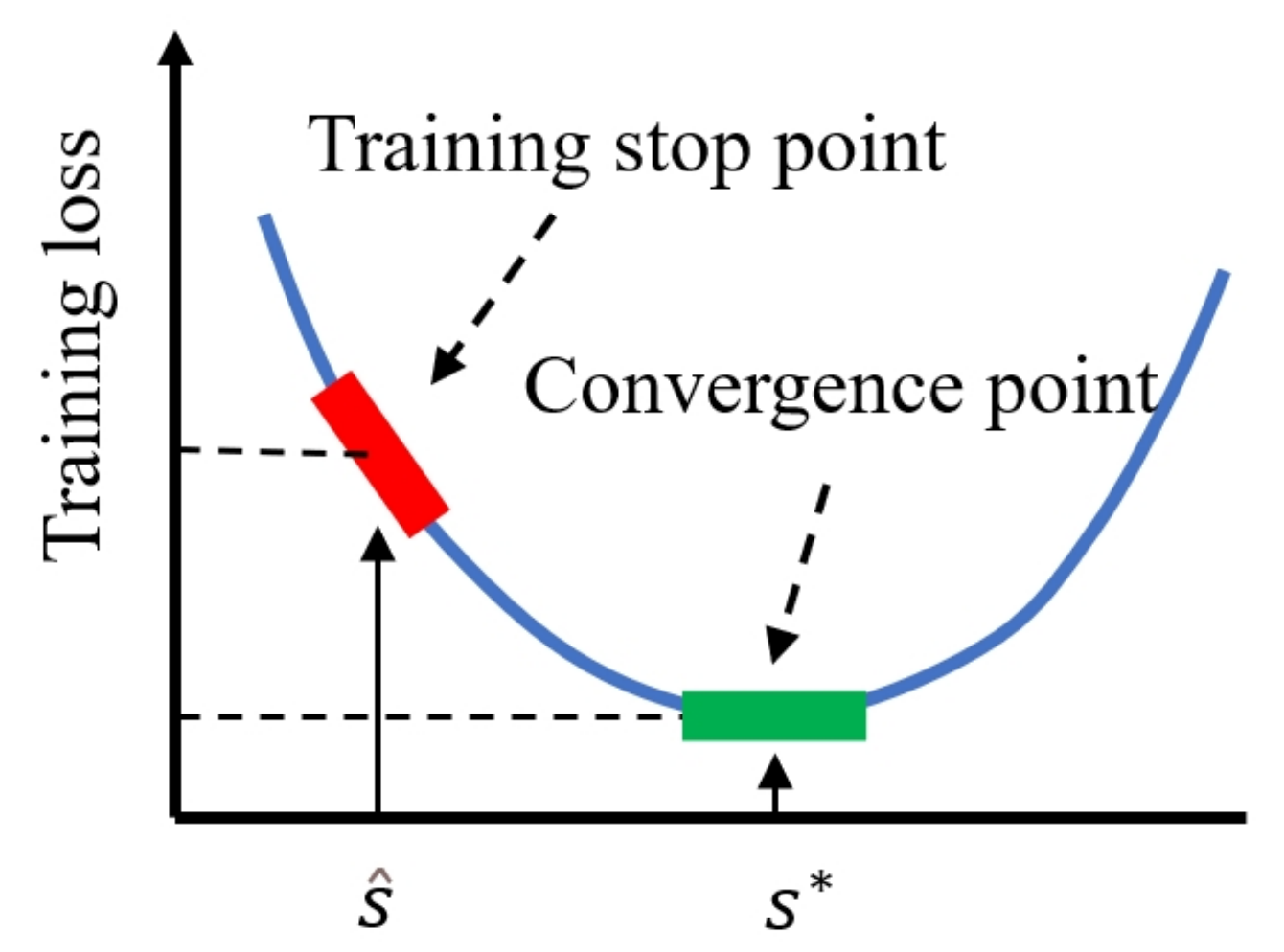}
    \caption{$\hat{s}$ may not converge to $s^*$ when training stops due to many factors such as insufficient samples, among other reasons.}
    \label{fig:training_loss}
\end{figure}

In scenarios with insufficient samples, it becomes more challenging to configure and adjust aspects such as the learning rate, initial weights, and batch size for neural networks. This makes it difficult for DRP's loss Equation (\ref{DRP}) to converge on the training samples, not to mention on the test samples during real-world deployment, even in the absence of covariate shift. Fig. \ref{fig:training_loss} illustrates this point.

This problem is common in the real world, too.
For example,
\begin{itemize}
    \item You are attempting to gather training samples for the DRP model from an online RCT to predict the ROI of carpooling coupons. If the RCT's treatment being tested negatively affect the user experience, it could result in reduced user engagement, further affecting long-term user retention and income. Hence, the RCT experiment can only utilize a minimal amount of online traffic and should 
    not be too long, which lead to the insufficient training data collection required for neural network convergence.
\end{itemize}



\subsection{Proposed Method}
\label{sec:c}

\begin{figure*}[!t]
    \centering
    \includegraphics[width=0.72\textwidth]{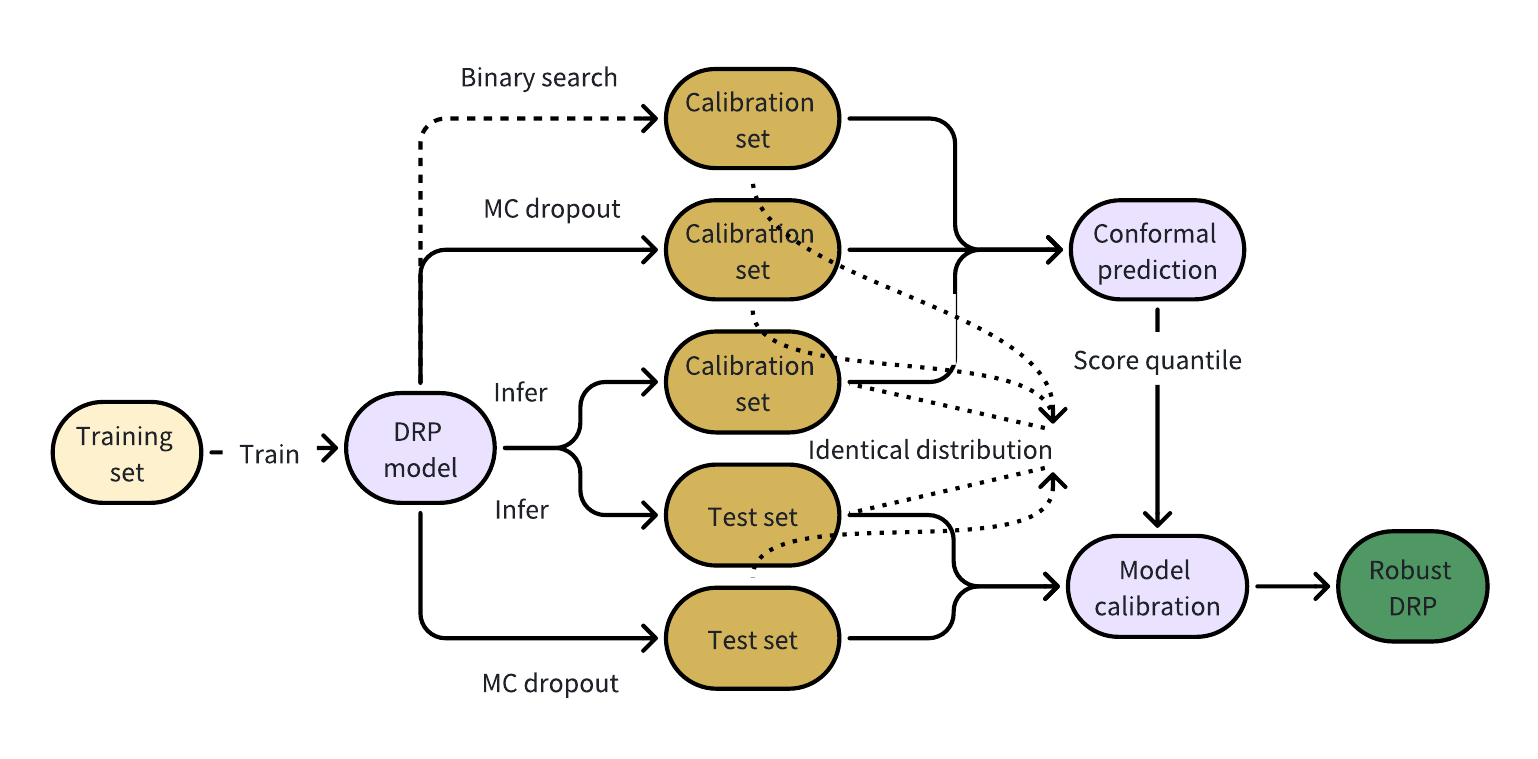}
    \caption{Architecture of rDRP.}
    \label{fig:arch}
\end{figure*}

\textbf{Robust DRP.} To mitigate the impact caused by the two aforementioned limitations, we propose a robust DRP method (rDRP). That is, adding a post-processing stage to the original DRP model, utilizing the information of the interval estimate to calibrate the point estimate results of DRP. The architecture of rDRP is illustrated in Fig. \ref{fig:arch}.

\textbf{Conformal Score.} We use conformal prediction to obtain rigorous prediction intervals for ROI. A popular method is Conformalized Quantile Regression (CQR) \cite{romano2019conformalized}. CQR combines quantile regression with conformal prediction, using the quantiles generated by quantile regression as the basis for the conformal prediction process to provide statistically guaranteed prediction intervals. However, CQR requires a quantile loss for the quantile regression. In this study, it is challenging to rewrite the convex loss function of DRP (see Equation (\ref{DRP})) as a quantile loss, and therefore, we are unable to use CQR. 

An alternative is by using \textit{Conformalizing Scalar Uncertainty Estimates} \cite{angelopoulos2021gentle}, which is easy to use and does not require any changes to the model structure or loss function of DRP, although we acknowledge it has some limitations (for a detailed discussion on the limitations, see section \ref{sec:dis}). 

Before applying it to our framework, we make two assumptions:
\begin{assumption}
\label{assu:1}
We assume that the convergence point of the convex loss function in DRP can be approximately considered as the true value of the ROI.
\end{assumption}

The "Conformalizing Scalar Uncertainty Estimates" \cite{angelopoulos2021gentle} requires the provision of the true value corresponding to the predicted value. However, for quantities such as ROI or uplift, it is impossible to obtain the true value in actual business scenarios. Nevertheless, the convex property of the DRP loss function allows us to obtain its convergence point, and approximately consider the value corresponding to the convergence point as the true value.

\begin{assumption}
\label{assu:2}
We assume that the calibration set and the test set have the same data distribution. 
\end{assumption}

This assumption ensures the strict interval coverage guarantee of conformal prediction (see Equation (\ref{eq:contain})) and also ensures that the conformal score calculated from the calibration set using "Conformalizing Scalar Uncertainty Estimates" (see Equation (\ref{eq:uncertainty_scalar_score})) can be directly used to calibrate the prediction results of the test set (see Equations \ref{eq:combine1} to \ref{eq:combine3}).

Therefore, we define the Conformal Score as follows: 
\color{black}

\begin{equation}
\label{eq:uncertainty_scalar_score}
    score(x,roi^*) = \frac{\left|roi^* - \hat{roi}\right|}{\hat{r}(x)},
\end{equation}
where $roi^*$ equals $\sigma(s^*)$ when $s^*$ is the convergence point, $\hat{roi}$ is actual prediction point of DRP and $\hat{r}(x)$ is the standard deviation of $\hat{roi}$. Note that in the process of calculating the Conformal Score, all three terms on the right side of the Equation (\ref{eq:uncertainty_scalar_score}) are conducted on the calibration set.

This part proceeds as follows: subsection \ref{sec:rDRP1} covers how to obtain $roi^*$ by binary search in Algorithm \ref{alg:binary}. Subsection \ref{sec:rDRP2} describes how to calculate $\hat{r}(x)$ through Monte Carlo Dropout. Subsection \ref{sec:rDRP3} covers how to get the rigorous prediction interval by following conformal prediction procedures in Algorithm \ref{alg:conformal}. With the rigorous prediction interval in hand, the subsection \ref{sec:rDRP4} introduces several heuristic method to calibrate the point estimate of DRP. The last subsection \ref{sec:rDRP5} combines all the above together to form the rDRP method in Algorithm \ref{alg:deepcr}.

\subsubsection{Obtain $roi^*$}
\label{sec:rDRP1}

DRP's loss Equation \eqref{DRP} is proved to be convex with respect to $s_i$ according to \textit{Theorem 2} in \cite{zhou2023direct}. Hence, to obtain the convergence point, we can just search for the global minimum point instead. A binary search is proposed in Algorithm \ref{alg:binary}.

\begin{algorithm}[ht!]
\caption{Obtain $roi^*$ with Binary Search}
\label{alg:binary}
\textbf{Input}: $(t_i, y^r_i, y^c_i)$ from calibration set\\
\textbf{Output}: $roi^*$
\begin{algorithmic}[1] 
\STATE Let $roi_l = 0, roi_r = 1, roi^* = \frac{roi_l + roi_r}{2}$.
\STATE Compute the derivative  $L'(s_i^*)$ at $s_i^* = \sigma^{-1}(roi^*)$.
\STATE Let $\epsilon$ be a small positive constant.

\WHILE{$\left|roi_r - roi_l\right|  > \epsilon$} 
\IF{$\left|L'(s_i^*)\right| < \epsilon$}
    \STATE break
 \ENDIF   
\IF{$L'(s_i^*) > 0$}
    \STATE $roi_r = roi^*$
\ELSE
    \STATE $roi_l = roi^*$
\ENDIF
    \STATE Let $roi^* = \frac{roi_l + roi_r}{2}$.
    \STATE Compute the derivative  $L'(s_i^*)$ at $s_i^* = \sigma^{-1}(roi^*)$.
\ENDWHILE
\end{algorithmic}
\end{algorithm}

\subsubsection{Calculate $\hat{r}(x)$}
\label{sec:rDRP2}

In neural network models, standard deviation (std) generation commonly involves ensemble methods or input perturbation variance, but these require retraining multiple models, which is inefficient. To circumvent these issues, we suggest using the Monte Carlo dropout method, which requires no changes to the model's training architecture and is applied only during the inference phase.

Monte Carlo Dropout, as introduced by \cite{gal2016dropout}, interprets regular dropout as a Bayesian approximation of Gaussian processes. This method allows for multiple predictions from a single input during inference by applying a retention probability $p$ to each neuron, creating different active neuron sets for each prediction. With repeated inputs under varying neuron activations, we obtain multiple point estimates $\hat{roi}$, enabling straightforward calculation of the standard deviation $\hat{r}(x)$.

\subsubsection{Conformal Prediction}
\label{sec:rDRP3}
By following conformal prediction procedures, in Algorithm \ref{alg:conformal}, we show how to calculate a rigorous prediction interval $C(x)$ and score quantile $\hat{q}$.
\begin{algorithm}[tb]
\caption{Calculate the Rigorous Prediction Interval $C(x)$ and Score Quantile $\hat{q}$}
\label{alg:conformal}
\textbf{Input}: $(roi^*, \hat{roi}, \hat{r}(x))$\\
\textbf{Output}: $C(x)$, $\hat{q}$
\begin{algorithmic}[1] 
\STATE Calculate $score(x,roi^*)$ by Equation (\ref{eq:uncertainty_scalar_score}).
\STATE Let $n$ be the size of the calibration set.
\STATE Arrange $n$ samples in increasing order by $score(x,roi^*)$.
\STATE Let $\alpha \in (0,1)$ be a user-chosen error rate.
\STATE Take $\hat{q}$ to be the $\frac{\lceil (1-\alpha)(n+1) \rceil}{n}$ quantile of $score(x,roi^*)$.
\STATE $C(x)=\left[ \hat{roi} - \hat{r}(x)\hat{q}, \hat{roi} + \hat{r}(x)\hat{q} \right]$.
\end{algorithmic}
\end{algorithm}
The conformal prediction guarantee that, for a new test data, the following holds:

\begin{equation}
\label{eq:contain}
P\Big(roi^*_{\rm test} \in C(x_{test})\Big) \ge 1 - \alpha,
\end{equation}
where $roi^*_{\rm test}$ is at the converge point of the test set and $\alpha$ is the error rate illustrated in Algorithm \ref{alg:conformal}. If $\alpha$ is set to be 0.1, with at least 90\% probability, we guarantee that $C(x_{test})$ contains the converge point $roi^*_{\rm test}$. \textit{Appendix A} in \cite{angelopoulos2021gentle} provides a detailed proof for the above Equation (\ref{eq:contain}).

We do not discuss the details of the proof here. However, we would like to emphasize that the proof is contingent upon certain prerequisite conditions, that is, the calibration set and test set share the same distribution. Next, we clarify why this prerequisite condition is met for Equation (\ref{eq:contain}) to hold in this study. Before each deployment of the model, we first conduct a very small-scale RCT in the real environment for one or two days. We then use this portion of RCT data as the model's calibration set, which is highly likely to have the same probability distribution as the test set we will face. This way, even if the distribution of the model's training set and the real-world test data are inconsistent, it does not affect our use of the conformal prediction method.



\subsubsection{Model Calibration}
\label{sec:rDRP4}
In order to get better performance for ROI's prediction, our next focus is how to calibrate the point estimates with conformal prediction interval.

\textit{The 2018 M4 Forecasting Competition} by Kaggle was the first M-Competition to elicit
prediction intervals in addition to point estimates. \cite{Grushka-Cockayne2020-qv} summarized six effective interval aggregation methods used in the competition. Although the goals are somewhat different, the competition's focus is on time series, while our focus is on ROI ranking sequence. This competition provided us with many insights and inspirations, hence we propose the following heuristic methods:
\begin{subequations} 
\begin{align}
\label{eq:combine1}
\widetilde{roi} & = \hat{roi}(\hat{roi} + \hat{r}(x)\hat{q}), \\
\widetilde{roi} & = \frac{\hat{roi}}{\hat{r}(x)\hat{q}}, \\
\label{eq:combine3}
\widetilde{roi} & = \hat{roi} + \hat{r}(x)\hat{q}, 
\end{align}
\end{subequations}
To select which equation needs to be validated on the validation/calibration set.

\subsubsection{rDRP}
\label{sec:rDRP5}

We propose a robust DRP (rDRP) method by combining all the above parts together.
The procedure to perform rDRP is summarized in Algorithm \ref{alg:deepcr}. Finally, with $\widetilde{roi}(x_{test})$ in hand, C-BTAP can be solved by Algorithm \ref{alg:BTAG}.

\begin{algorithm}[tb]
\caption{Robust DRP (rDRP)}
\label{alg:deepcr}
\textbf{Input}: Training, calibration and test RCT samples.\\
\textbf{Output}: $\widetilde{roi}(x_{test})$
\begin{algorithmic}[1] 
\STATE \textbf{First, on the training set:}
\STATE \qquad (i) Train DRP model. \label{alg:step1}
\STATE \textbf{Second, on the calibration set:}
\STATE \qquad (i) Infer DRP model to obtain $\hat{roi}$. \label{alg:step4}
\STATE \qquad (ii) Calculate $roi^*$ by Algorithm \ref{alg:binary}. \label{alg:step5}
\STATE \qquad (iii) Infer DRP's MC Dropout model to obtain $\hat{r}(x)$. \label{alg:step6}
\STATE \qquad (iv) Calculate $\hat{q}$ by Algorithm \ref{alg:conformal}. \label{alg:step7}
\STATE \qquad (v) Select $\widetilde{roi}$'s calibration form from \ref{eq:combine1} to \ref{eq:combine3}. \label{alg:step8}
\STATE \textbf{Third, on the test set:}
\STATE \qquad (i) Infer DRP model to obtain $\hat{roi}$. \label{alg:step10}
\STATE \qquad (ii) Infer DRP's MC Dropout to obtain $\hat{r}(x)$. \label{alg:step11}
\STATE \qquad (iii) With $\hat{q}$ in line \ref{alg:step7} and the selected form in line \ref{alg:step8}, $\widetilde{roi}(x_{test})$ is obtained. \label{alg:step12}
\end{algorithmic}
\end{algorithm}

\subsection{Time Complexity Analysis For Algorithm \ref{alg:deepcr}}
\label{sec:d}

In this subsection, we analyze the time complexity of our proposed rDRP compared to the original DRP. According to Algorithm \ref{alg:deepcr} of rDRP, this analysis can be divided into three parts for discussion.

\begin{enumerate}
    
    \item \textbf{Training phase.} During the model training phase, rDRP directly uses the DRP model, so the time complexity is the same compared to the DRP model (line \ref{alg:step1} in Algorithm \ref{alg:deepcr}).

    \item \textbf{Calibration phase.} In line \ref{alg:step4}, suppose the DRP inference time for each sample is $\Delta_{infer}$. The value of $\Delta_{infer}$ will be very small, considering that the neural network of DRP includes only one hidden layer, with the number of nodes in this hidden layer ranging between 10 to 100 in this study. In line \ref{alg:step5}, each sample requires executing a binary search of Algorithm \ref{alg:binary} once, and the time complexity of performing a binary search once is at most $\lfloor \log_2{(\frac{1}{\epsilon})}\rfloor + 1$ iterations, where $\epsilon$ usually is a decimal, like 0.001. In line \ref{alg:step6}, MC dropout needs to execute multiple times DRP's inference, usually 10 to 100 times. Suppose  the calibration size is $N_{cali}$, which is usually 1000 to 10000. For each sample, denote the total time complexity from line \ref{alg:step4} to \ref{alg:step6} as a constant $k$, which is not related to $N_{cali}$. So the total time complexity up to line \ref{alg:step6} is $O(kN_{cali})$. In line \ref{alg:step7}, searching the quantile is the main part of time complexity, and the time complexity of quantile searching is $O(N_{cali}log(N_{cali}))$. In line \ref{alg:step8}, a simple grid search can be conducted, whose time complexity can be neglected. To sum up the calibration phase, the total time complexity is $O(N_{cali}(k + log(N_{cali})))$, which is acceptable in most industrial scenarios, especially considering that $N$ is not particularly large, and the calibration phase can be completed in advance offline. Note, DRP does not need this phase.
    
    \item \textbf{Inference phase.} In line \ref{alg:step10}, the time complexity is $\Delta_{infer}$. In line \ref{alg:step11}, time complexity is 10 to 100 times $\Delta_{infer}$. However, line \ref{alg:step11} can be executed in parallel if we have enough computation resource. The time complexity of line \ref{alg:step12} can be neglected. Note, in this phase, DRP only needs one $\Delta_{infer}$.

\end{enumerate}    
In a word, during the Training phase, DRP and rDRP have the same time complexity. In the Calibration phase, DRP does not require this phase, while the time complexity of rDRP is $O(N_{cali}(k + log(N_{cali})))$. However, this phase can be conducted offline in advance, and the value of $N_{cali}$ is generally not large. During the Inference phase, DRP only requires one inference, but rDRP requires multiple inferences. However, if sufficient computational resources are available, the inferences for rDRP can be executed in parallel and almost with the same time delay as DRP.

\color{black}
\section{Experiments}
\label{sec:experiment}
To compare the performance of rDRP with other state-of-the-arts, extensive offline and online tests are conducted on real-world datasets.

\subsection{Offline Test}
\label{sec:offline}


\begin{table*}[!t]
\centering
\caption{For three public real-world dataset {CRITEO-UPLIFT v2}, {Meituan-LIFT} and {Alibaba-LIFT}, offline AUCC evaluation results in four settings: \textbf{SuNo}, \textbf{SuCo}, \textbf{InNo} and \textbf{InCo}, respectively.}
\label{table:1}
\begin{tabular}{c|c|cc|cc|cc}
\toprule
\multirow{2}{*}{Dataset Size} & \multirow{2}{*}{Method} & \multicolumn{2}{c}{{CRITEO-UPLIFT v2}} 
& \multicolumn{2}{c}{{Meituan-LIFT}}
& \multicolumn{2}{c}{{Alibaba-LIFT}}\\ 
\cmidrule(lr){3-4} 
\cmidrule(lr){5-6}
\cmidrule(lr){7-8}
& & \pmb{No} Covariate Shift  &  \pmb{Co}variate Shift 
& \pmb{No} Covariate Shift  &  \pmb{Co}variate Shift
& \pmb{No} Covariate Shift  &  \pmb{Co}variate Shift\\
\midrule                            
 \multirow{3}{*}{ \pmb{Su}fficient}

&TPM-SL        &$0.6983$  &{$0.6824$} &0.6890 &0.5938 &0.7213  &0.6975 \\ 
&TPM-XL        &$0.5965$  &{$0.6108$} &0.7213 &0.6494 &0.7234  &0.6950 \\ 
&TPM-CF        &$0.7034$  &{$0.6817$} &0.5841 &0.5202 &0.7177  &0.6241 \\ 
&TPM-DragonNet        &0.6497  &0.6712 &0.5478 &0.5844 &0.7079  &0.6846 \\ 
&TPM-TARNet         &0.7359  &0.6500 &0.5147 &0.5683 &0.7264  &0.6509 \\ 
&TPM-OffsetNet        &0.7115  &0.5433  &0.5164 &0.5038 &0.7275  &0.6215 \\ 
&TPM-SNet        &0.6953  &0.6411 &0.5392 &0.4766 &0.6392  &0.6390 \\ 
&DR        &$0.7474$  &{$0.6757$} &0.6067 &0.6421 &0.6214  &0.5422 \\ 
&DRP        &$0.7714$  &{$0.7263$}  &0.7223 &0.6580 &0.7281  &0.6867 \\ 
&\pmb{rDRP}  &\pmb{$0.7717$} &\pmb{$0.7382$ } &\pmb{0.7290} &\pmb{0.6611} &\pmb{0.7476}  &\pmb{0.7042} \\

\midrule
\multirow{3}{*}{ \pmb{In}sufficient}

&TPM-SL        &$0.5772$  &{$0.5851$} &0.6248 &0.5747 &0.7082  &0.6204 \\ 
&TPM-XL        &$0.5797$  &{$0.4215$} &0.6494 &0.5807 &0.7035  &0.6541 \\ 
&TPM-CF        &$0.5875$  &{$0.5358$} &0.5935 &0.5720 &0.6134  &0.6518 \\ 
&TPM-DragonNet &0.6203 &0.5374 &0.6118 &0.5807 &0.6998  &0.6402 \\
&TPM-TARNet &0.6190 &0.5371 &0.6959 &0.5646 &0.6570  &0.6360 \\
&TPM-OffsetNet &0.5373 &0.5196 &0.6088 &0.6692 &0.6651  &0.6366 \\
&TPM-SNet &0.6287 &0.5504 &0.6209 &0.6210 &0.6686  &0.6637 \\
&DR        &$0.6155$  &{$0.4465$} &0.6041 &0.5736 &0.5888  &0.5888 \\ 
&DRP       &{$0.6222$}    &{$0.5411$} &0.6881 &0.6489 &0.7121  &0.6475 \\ 
&\pmb{rDRP} &\pmb{$0.6509$}   &\pmb{$0.6087$} &\pmb{0.7005} &\pmb{0.6753} &\pmb{0.7214}  &\pmb{0.6823} \\ 
                          
\bottomrule                           
\end{tabular}
\end{table*}

\textbf{Datasets.}
We evaluate our method on three public real-world industrial datasets: CRITEO-UPLIFT v2~\cite{Diemert2018}, Meituan-LIFT \cite{huang2024entire} and Alibaba-LIFT \cite{ke2021addressing}, respectively. 
\begin{itemize}
    \item
    \textbf{CRITEO-UPLIFT v2}. For a fair comparison, we use the same dataset as the one used to evaluate the DRP model in \cite{zhou2023direct}, that is, the CRITEO-UPLIFT v2 dataset, which is also one of the most widely recognized and evaluated datasets for uplift related models. The dataset was provided by Criteo for the AdKDD'18 workshop, as documented in ~\cite{Diemert2018}. Originating from a randomized controlled trial (RCT) designed to withhold advertising from a randomly selected subset of users, it comprises 12 feature variables, a singular binary indicator for treatment, and two outcome labels (visit and conversion). For the purpose of evaluating various models on their ability to predict individual ROI, the visit outcome is treated as the cost factor, while the conversion outcome represents the benefit. The dataset
    encompasses a total of 13.9 million data points.
    \item 
    
    \textbf{Meituan-LIFT}. This dataset \cite{huang2024entire} originates from a two-month randomized controlled trial (RCT) focused on smart coupon marketing for food delivery within the Meituan app, a leading platform for local services in China. It encompasses nearly 5.5 million entries, featuring 99 attributes, detailed intervention data, and two outcomes: clicks and conversions. The treatment variable is multifaceted, offering five distinct options. In the context of assessing C-BTAP, clicks are analyzed as a cost element, whereas conversions signify the benefit. From the five available treatment options, only two are chosen for consideration, with these selected interventions being simplified into a binary treatment format for analysis.
    \item
    \textbf{Alibaba-LIFT}. It is an RCT dataset of uplift modeling for different brands in a large-scale advertising scene, which is open-sourced by Alibaba \cite{ke2021addressing}. This dataset comprises billions of instances, incorporating twenty-five discrete features and nine multivalued features, along with binary treatments and two outcome labels (exposure and conversion). For the purpose of evaluating various models on their ability to predict individual ROI, the exposure outcome is treated as the cost factor, while the conversion outcome represents the benefit.
\end{itemize}

\textbf{Settings.}
Based on whether there are sufficient samples and whether there is covariant shift between the calibration/test and training sets, we consider four different settings. Note that the insufficient dataset are randomly taken from the sufficient dataset with a 0.15 sample rate. The covariant shift between the calibration/test and training sets is achieved by altering the distribution of the features only in the calibration and test sets, that is, those features in the training set remain unchanged.
\begin{itemize}
    \item
    \textbf{Su}fficient data and \textbf{No} covariate shift (\textbf{SuNo}). 
    \item
    \textbf{Su}fficient data and \textbf{Co}variate shift (\textbf{SuCo}). 
    \item
    \textbf{In}sufficient data and \textbf{No} covariate shift (\textbf{InNo}). 
    \item
    \textbf{In}sufficient data and \textbf{Co}variate shift (\textbf{InCo}).
\end{itemize}

\textbf{Evaluation Metric.} We use the same metric as evaluating DRP in \cite{zhou2023direct}.
\begin{itemize}
    \item Area under Cost Curve (\textbf{AUCC}). A commonly used metric for evaluating the performance to rank ROI of individuals~\cite{Du2019Improve}. It is calculated by plotting a curve that represents the cumulative cost against the cumulative benefit for each decile or percentile of the data. In simple terms, it first sorts data by predicted ROI, then calculate the cumulative benefit and cumulative cost at each percentile, and plot this cumulative curve. The area under this curve is then computed, with a larger area indicating a more cost-effective model.
\end{itemize}

\textbf{Benchmark.} For each of the four scenarios mentioned in the aforementioned settings, we compare the effectiveness of the following ten methods for C-BTAP. 

\begin{itemize}
    \item Cost-aware Binary treatment assignment problem (C-BTAP)
        \begin{itemize}    
            \item \textbf{TPM} (Including seven models). The Two Phase Method (TPM) utilizes two uplift models to separately estimate incremental revenue and cost. It then calculates an individual's ROI by dividing the revenue prediction by the cost prediction. We select a set of the most representative uplift methods, including S-Learner (SL) \cite{kunzel2019metalearners}, X-Learner (XL) \cite{kunzel2019metalearners}, Causal Forest (CF) \cite{Athey2019Generalized}, DragonNet \cite{shi2019adapting}, TARNet\cite{shalit2017estimating}, OffsetNet \cite{curth2021inductive}, and SNet \cite{curth2021nonparametric}. In our study, these baselines are sequentially denoted as \textbf{TPM-SL}, \textbf{TPM-XL}, \textbf{TPM-CF}, \textbf{TPM-DragonNet}, \textbf{TPM-TARNet}, \textbf{TPM-OffsetNet} and \textbf{TPM-SNet}, respectively.

            \item \textbf{DR}. In the model of Direct Rank (DR), a loss function aimed at ranking individuals' ROI is similarly created, as noted in ~\cite{Du2019Improve}. However,  \cite{zhou2023direct} demonstrate that achieving accurate ranking is not possible when the loss function fully converges, as detailed in Appendix E of \cite{zhou2023direct}.

            \item \textbf{DRP}. Proposed by \cite{zhou2023direct} in \textit{AAAI 2023}, DRP is a Direct ROI Prediction model for Cost-aware Binary Treatment Assignment Problem (BTAP). Based on our research of the published literature, the DRP remains the State-Of-The-Art (SOTA) model for C-BTAP so far.

            \item \textbf{rDRP}. A robust DRP method proposed in this paper. In rDRP, the standard deviation (std) part produced by Monte Carlo (MC) dropout is calibrated using conformal Prediction (CP), that is, rDRP essentially equals to DRP with MC and CP method.
        \end{itemize}
\end{itemize}

For a fair comparison, the hyperparameters for training DRP and rDRP in this paper are the same and are consistent with those in \cite{zhou2023direct}. 


\textbf{Results.} For three public real-world dataset {CRITEO-UPLIFT v2}, {Meituan-LIFT} and {Alibaba-LIFT}, in Table \ref{table:1} we report the detailed offline AUCC evaluation results in four settings: {SuNo}, {SuCo}, {InNo} and {InCo}, respectively. We can see that rDRP achieves a larger AUCC than other methods, which proves the effectiveness of our method. The gap between DRP and rDRP is even more pronounced when both the dataset is insufficient and covariate shift exists. This is exactly the robustness of rDRP that we are hoping to see.




                          

\subsection{Ablation Study}


\begin{table*}[!t]
\centering
\caption{{Ablation study AUCC results: the contributions of MC to DR method and DRP method, respectively; the contributions of CP to DRP method.}}
\label{table:ablation_ts}
\begin{tabular}{c|c|cc|cc|cc}
\toprule
\multirow{2}{*}{Dataset Size} & \multirow{2}{*}{Method} & \multicolumn{2}{c}{{CRITEO-UPLIFT v2}} 
& \multicolumn{2}{c}{{Meituan-LIFT}}
& \multicolumn{2}{c}{{Alibaba-LIFT}}\\ 
\cmidrule(lr){3-4} 
\cmidrule(lr){5-6}
\cmidrule(lr){7-8}
& & \pmb{No} Covariate Shift  &  \pmb{Co}variate Shift 
& \pmb{No} Covariate Shift  &  \pmb{Co}variate Shift
& \pmb{No} Covariate Shift  &  \pmb{Co}variate Shift\\
\midrule                            
 \multirow{3}{*}{ \pmb{Su}fficient}

&DR        &$0.7459$  &{$0.6757$} &0.6067 &0.6421 &0.6214 &0.5422 \\ 
&DR w/ MC       &$0.7464$  &{$0.6988$} &0.6675 &0.6591 &0.6273 &0.5527 \\ 
&DRP        &$0.7714$  &{$0.7263$}  &0.7223 &0.6580 &0.7281 &0.6867 \\ 
&DRP w/ MC       &$0.7716$  &{$0.7265$}  &0.7253 &0.6596 &0.7393 &0.6938 \\ 
&DRP w/ MC w/ CP &\pmb{$0.7717$} &\pmb{$0.7382$ } &\pmb{0.7290} &\pmb{0.6611} &\pmb{0.7476} &\pmb{0.7042} \\

\midrule
\multirow{3}{*}{ \pmb{In}sufficient}

&DR        &$0.6155$  &{$0.4465$} &0.6041 &0.5736 &0.5914 &0.5888 \\ 
&DR w/ MC       &$0.6203$  &{$0.5326$} &0.6194 &0.6034 &0.6075 &0.6304 \\ 
&DRP       &{$0.6222$}    &{$0.5411$} &0.6881 &0.6489 &0.7121 &0.6475 \\ 
&DRP w/ MC      &{$0.6333$}  &{$0.5907$} &0.6935 &0.6609 &0.7166 &0.6746 \\ 
&DRP w/ MC w/ CP &\pmb{$0.6509$}   &\pmb{$0.6087$} &\pmb{0.7005} &\pmb{0.6753} &\pmb{0.7214} &\pmb{0.6823} \\

\bottomrule                           
\end{tabular}
\end{table*}

\begin{figure*}[!t]
\centering
\subfloat[]{\includegraphics[width=0.5\textwidth]{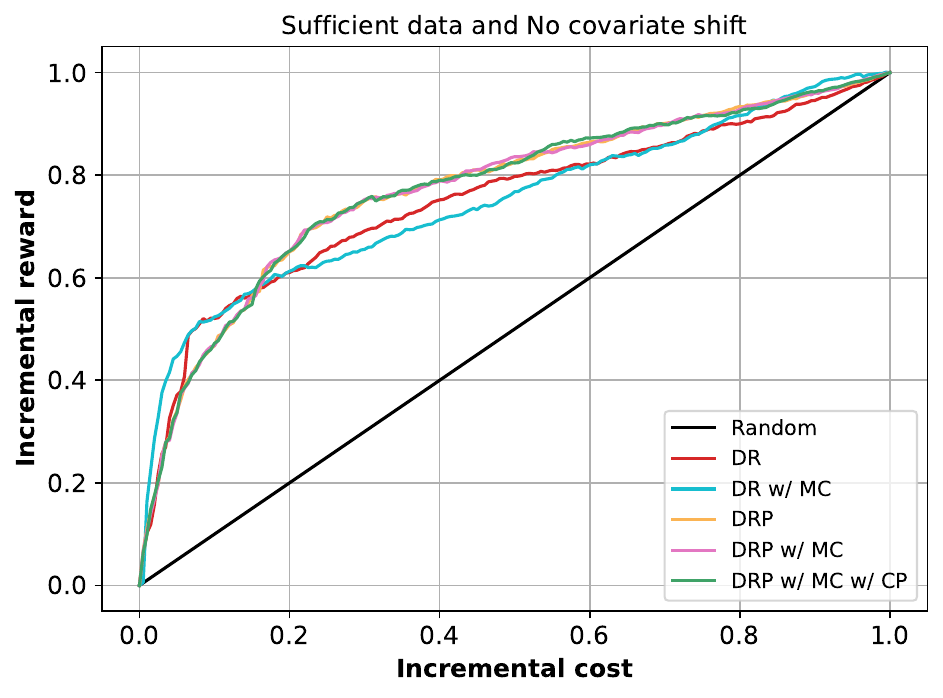}%
\label{fig:mae}}
\subfloat[]{\includegraphics[width=0.5\textwidth]{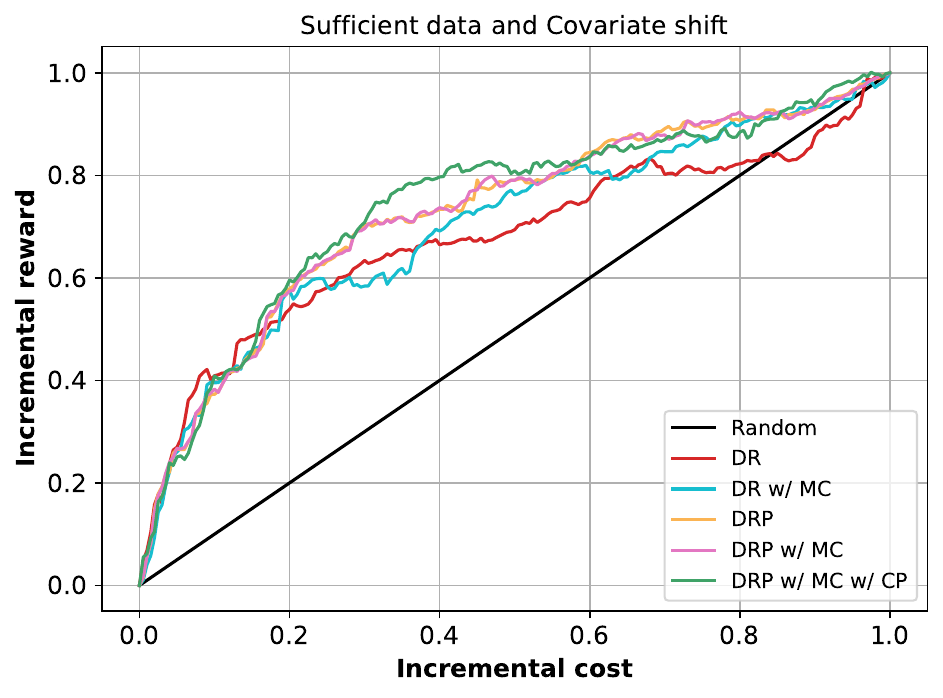}%
\label{fig:rmse}}
\hfil
\subfloat[]{\includegraphics[width=0.5\textwidth]{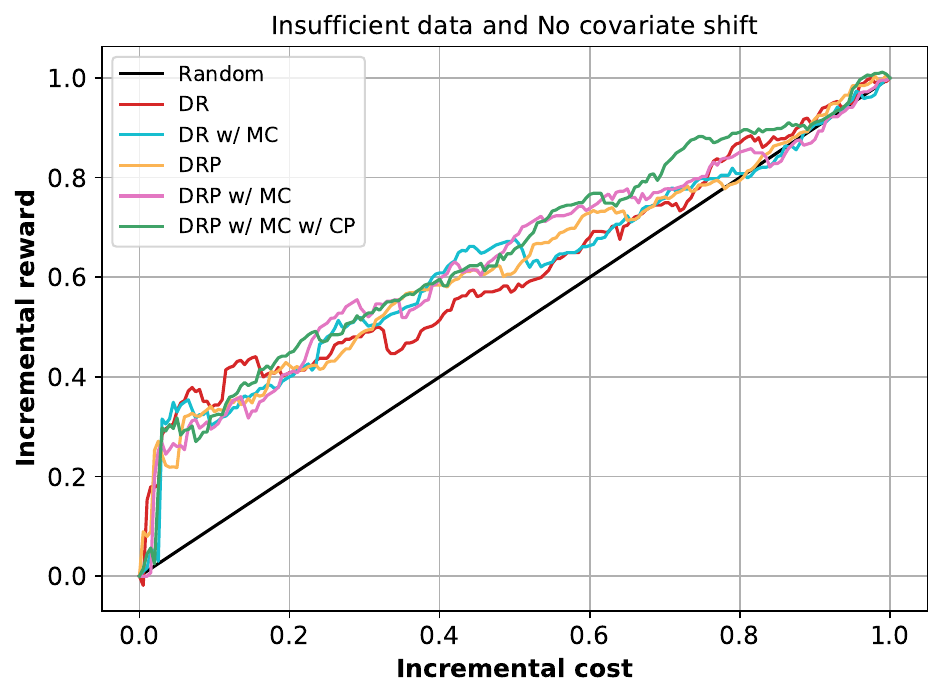}%
\label{fig:mape}}
\subfloat[]{\includegraphics[width=0.5\textwidth]{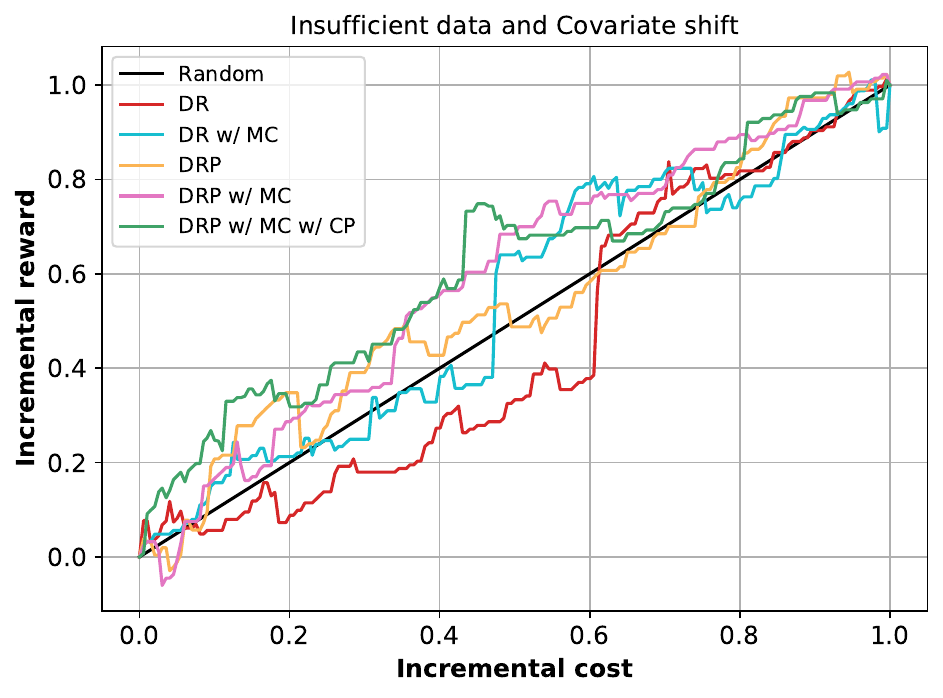}%
\label{fig:mape2}}

\caption{Ablation study AUCC results in four settings on dataset CRITEO-UPLIFT v2. (a) SuNo. (b) SuCo. (c) InNo. (d) InCo.}
\label{fig:offline_results}
\vspace{-10pt} 
\end{figure*}

\label{sec:abla}
To assess the effectiveness of each component (MC and CP) in rDRP, we conducted a series of ablation studies on the C-BTAP task. We use the same \textbf{datasets}, the same \textbf{settings} and the same \textbf{evaluation metric} as Offline Simulation in \ref{sec:offline}.

\begin{itemize}
    \item \textbf{Effect of Monte Carlo (MC) Dropout}.

    Note, the MC dropout method cannot be applied to TPM. The reason is as follows: TPM is composed of two uplift models, and using MC dropout for each uplift model can obtain the standard deviation (std) of each uplift. However, TPM also involves dividing one uplift by another to obtain the ROI, but the std of the one uplift divided by the std of another uplift does not yield the std of the predicted ROI. Only methods that directly predict ROI can use the MC approach, which in this study includes the DR and DRP methods.
        \begin{itemize}

            \item \textbf{DR w/ MC}. DR with MC dropout. During inference with the DR model, adding a Monte Carlo (MC) dropout layer enables the estimation of the standard deviation (std) for the DR point estimate. The DR w/ MC result is then derived by combining the DR's point estimate and std.
            
            \item \textbf{DRP w/ MC}. DRP with MC dropout. During inference with the DRP model, adding a Monte Carlo (MC) dropout layer enables the estimation of the standard deviation (std) for the DRP point estimate. The DRP w/ MC result is then derived by combining the DRP's point estimate and std.
        \end{itemize}  
        
    \item \textbf{Effect of Conformal Prediction (CP)}.

    Note, the CP calibration method can only be applied to DRP. The reasons are as follows: Firstly, the division form of TPM makes it impossible to directly use the MC dropout method to calculate the standard deviation (std), let alone apply CP calibration to the std. Secondly, for the DR method, since the loss function of DR is non-convex, it is not possible to use the binary search (Algorithm \ref{alg:binary}) to find the convergence point's ROI, thus CP calibration cannot be applied, as detailed in Equation (\ref{eq:uncertainty_scalar_score}). Therefore, we can only perform CP calibration on the std of DRP with MC.
        \begin{itemize}   
            \item \textbf{DRP w/ MC w/ CP}. DRP with MC and with CP. A robust DRP method proposed in this paper. In rDRP, the standard deviation (std) part produced by Monte Carlo (MC) dropout is calibrated using conformal Prediction (CP), that is, rDRP essentially equals to DRP w/ MC w/ CP method.
        \end{itemize}
\end{itemize}

\textbf{Results.} For three public real-world dataset {CRITEO-UPLIFT v2}, {Meituan-LIFT} and {Alibaba-LIFT}, in Table \ref{table:ablation_ts} we report the detailed offline AUCC evaluation results in four settings: {SuNo}, {SuCo}, {InNo} and {InCo}, respectively. From the results in Table \ref{table:ablation_ts}, we can observe that adding the component MC can result in a performance improvement for DR and DRP. Besides, with the component CP, the AUCC of DRP w/ MC can be further improved. In a word, this ablation study verifies the validity of the contributions MC and CP. These results for CRITEO-UPLIFT v2 dataset is also shown in Fig. \ref{fig:offline_results}.
\color{black}
\subsection{Online A/B Test}

We conduct four online A/B tests for our method in the incentivized advertising scenario on a short video platform. Incentivized advertising or rewarded ads, are a type of advertising where the viewer is rewarded for opting in to watch the ad. The form of rewards for viewers can include coins, cash, or online shopping vouchers, among others. In the context of incentivized advertising, based on our numerous past online experiments, a five-day A/B test is sufficient to yield reliable results for metrics such as ad revenue. The four tests correspond to the four settings mentioned in the offline simulation \ref{sec:offline}.

\textbf{Setups.} In each test, the viewers are randomly divided into three groups, that is, DRP, rDRP and Random Control, respectively. All these three groups are allocated the same reward budget. The difference in strategy lies in the differing ROI values predicted for these three groups: the first group through DRP, the second group through rDRP, and the third group through a random generator. The objective of the A/B tests is to maximize the advertising revenue accrued by the platform. 

The scenario for the four tests are described as follows:
\begin{itemize}
    \item
    \textbf{Su}fficient data and \textbf{No} covariate shift (\textbf{SuNo}). There are 15 million samples for training DRP or rDRP model. The training samples are collected during workday period and the trained model is set to be deployed online during the forthcoming workday.
    \item
    \textbf{Su}fficient data and \textbf{Co}variate shift (\textbf{SuCo}). There are 15 million samples for training DRP or rDRP model. The training samples are collected during workday period but the trained model is set to be deployed online during the forthcoming holiday or marketing campaign.
    \item
    \textbf{In}sufficient data and \textbf{No} covariate shift (\textbf{InNo}). There are 1.5 million samples for training DRP or rDRP model. The training samples are collected during workday period and the trained model is set to be deployed online during the forthcoming workday.
    \item
    \textbf{In}sufficient data and \textbf{Co}variate shift (\textbf{InCo}). There are 1.5 million samples for training DRP or rDRP model. Moreover, the training samples are collected during workday period but the trained model is set to be deployed online during the forthcoming holiday or marketing campaign.
\end{itemize}

\begin{figure*}[!t]
\centering
\subfloat[]{\includegraphics[width=0.24\textwidth]{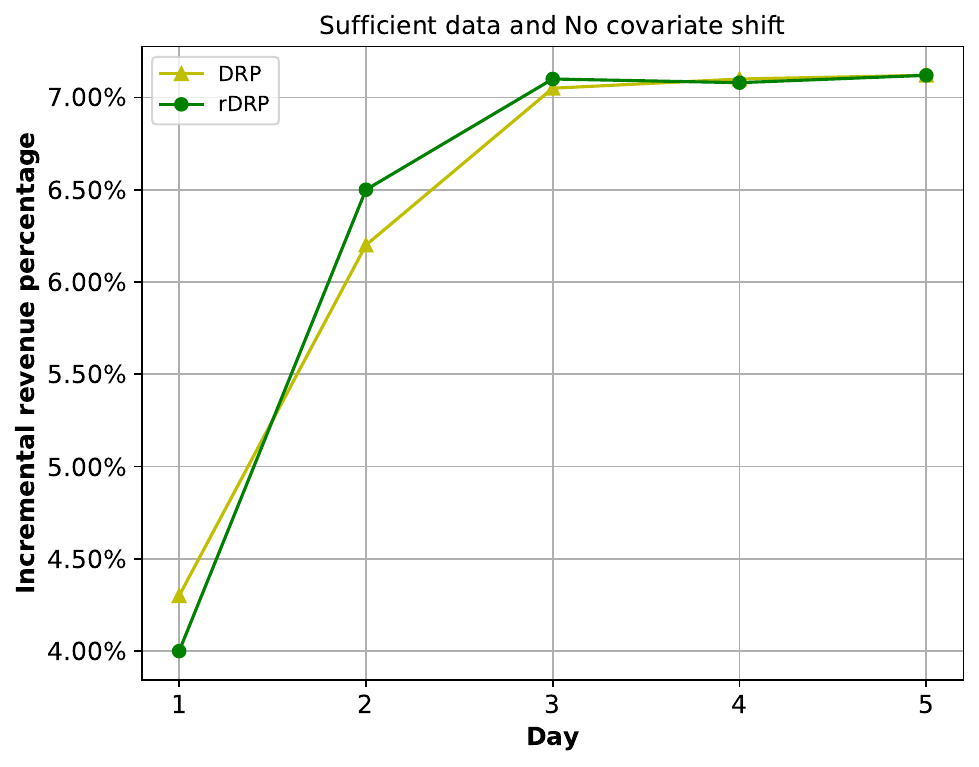}%
\label{fig:mae}}
\subfloat[]{\includegraphics[width=0.24\textwidth]{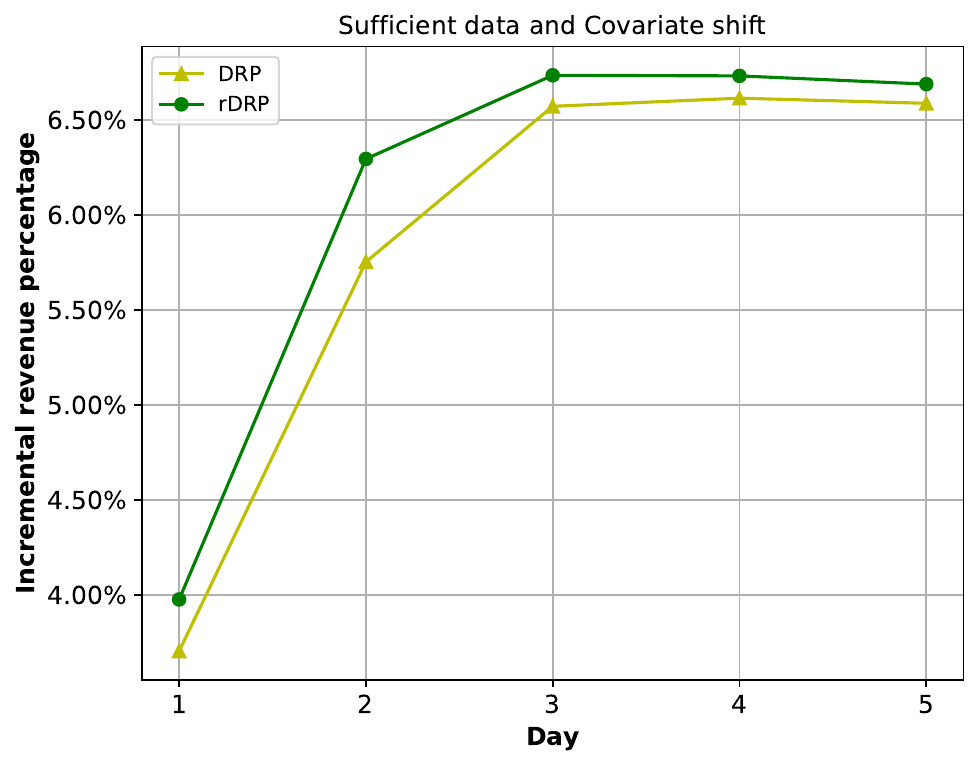}%
\label{fig:rmse}}
\subfloat[]{\includegraphics[width=0.24\textwidth]{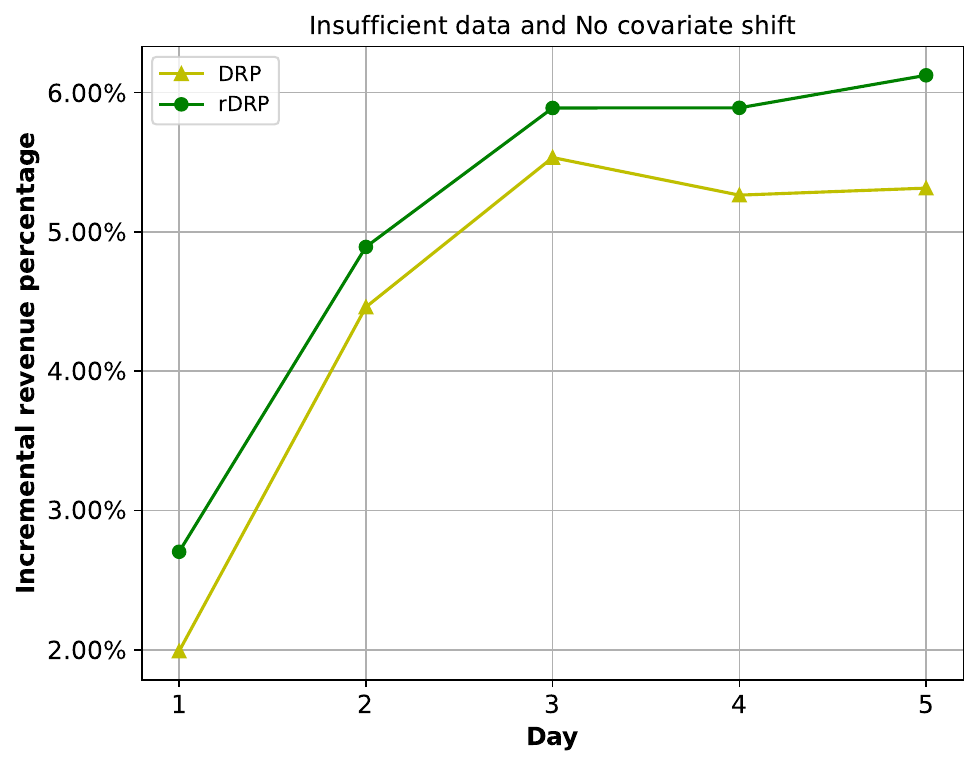}%
\label{fig:mape}}
\subfloat[]{\includegraphics[width=0.24\textwidth]{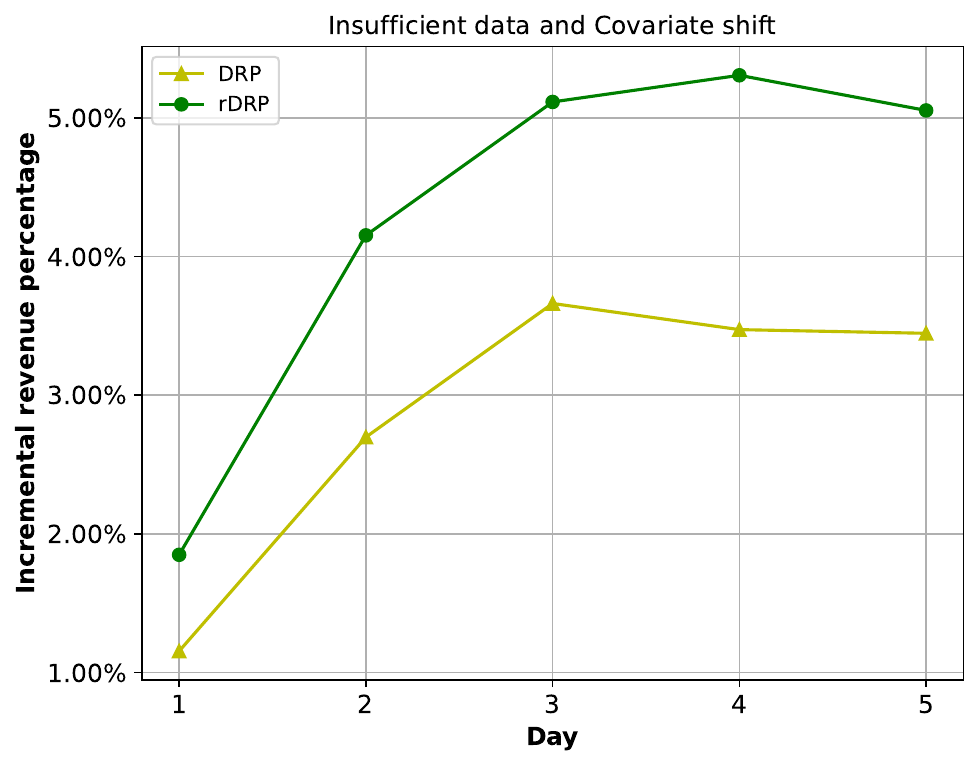}%
\label{fig:mape2}}

\caption{Online A/B test results in four settings. (a) SuNo. (b) SuCo. (c) InNo. (d) InCo.}
\label{fig:online_results}
\vspace{-10pt} 
\end{figure*}

\textbf{Results.} Fig. \ref{fig:online_results} show that, compared to the random control group, the percentage increase in advertising revenue for DRP and rDRP, respectively. The conclusions are essentially consistent with the offline evaluation. Except for SuNo, in the other three scenarios, rDRP shows significant improvement.


\section{Discussion}
\label{sec:dis}

In this section, we discuss the following five limitations of our proposed rDRP, especially in practical industrial scenarios. 

\begin{itemize}
    \item \textbf{Binary Treatment.} The rDRP focuses on binary treatment, meaning the treatments represent the presence or absence of the intervention, respectively. rDRP cannot be directly applied to multiple treatments. Multiple treatments refer to interventions or conditions where more than two levels or states are possible. Multiple treatments can involve comparing several different drugs, various doses of a single drug, different educational approaches, policy variations, and so on.
    However, Divide and Conquer method can be adopted for multiple treatment, which decomposes the multiple treatment problem into several binary treatment problems. Then each binary treatment problem can use the rDRP method proposed in this study.
    
    \item \textbf{RCT Samples.} Like DRP, rDRP can only be trained on RCT samples and cannot be trained on non-RCT samples such as observational datasets. 

    \item \textbf{Run Time Consideration.} As analyzed in the subsection \ref{sec:d}, during the actual online deployment phase of inference, rDRP requires multiple inferences for a single sample. If there is not enough computational resource, such as GPUs, it might lead to online latency in inference. Regarding this, careful consideration is needed to balance the following two factors: the cost burden of adding machine resources and the business benefits brought by adopting the rDRP method.

    \item \textbf{Positive Treatment Effect.} Like DRP, rDRP can only be applied to the scenario where marketing interventions positively influence an individual's response and also entail a positive cost, that is, $\tau^r(x_i) > 0$ and $\tau^c(x_i) > 0$.

    \item \textbf{Conformalizing Scalar Uncertainty Estimates.} A key component in Conformalizing Scalar Uncertainty Estimates is the uncertainty scalar, such as the std $\hat{r}(x)$ in Equation (\ref{eq:uncertainty_scalar_score}). However, as verified by \cite{angelopoulos2022image}, uncertainty scalars, which are used to adjust the width of the prediction interval, may not scale appropriately with the error rate $\alpha$ in the Equation (\ref{eq:contain}). As a result, essentially, increasing or decreasing $\alpha$ might not proportionately adjust the length of the prediction interval as one might expect. This implies a potential disconnect between the theoretical basis of uncertainty quantification and its practical application.
\end{itemize}
\color{black}

\section{Conclusion}
\label{sec:conclusion}
This work addresses the challenge of resource allocation in various commercial sectors through intelligent decision-making using data mining and neural network technologies, focusing on ROI optimization. We examine the Cost-aware Binary Treatment Assignment Problem (C-BTAP) and the SOTA Direct ROI Prediction (DRP) method, highlighting its limitations like covariate shift and insufficient training data. To overcome these, we propose the robust DRP method, utilizing conformal prediction for interval estimation and a heuristic calibration approach inspired by the M4 Forecasting Competition. The effectiveness of our methods is validated through offline simulations and online A/B tests, demonstrating significant improvements over existing models when facing insufficient data or covariate shift.

For the future work, first we will focus on how to rigorously estimate intervals for neural network-based uplift models on non-RCT datasets, i.e., observational datasets. Second, in multiple treatment scenarios, if not using the divide and conquer method that combines multiple rDRPs, is it possible to directly estimate their statistically rigorous prediction intervals? Third, for the method of calibrating point estimates of uplift models using interval estimation information, is there a more reasonable and rigorous approach than the current heuristic methods, etc. These are all directions that can be considered in the future and are also very challenging topics.

\bibliographystyle{IEEEtran}
\bibliography{IEEEabrv,refs}
\end{document}